%% file: main.tex
\documentclass[sigconf,nonacm]{acmart}

\usepackage{booktabs}
\usepackage{array}
\usepackage{graphicx}
\usepackage{afterpage}
\setcopyright{acmlicensed}
\copyrightyear{2027}
\acmYear{2027}
\acmDOI{XXXXXXX.XXXXXXX}
\acmConference[CHI '27]{CHI Conference on Human Factors in Computing
  Systems}{TBD}{TBD}
\acmISBN{978-1-4503-XXXX-X/27/XX}

\title[Calibrated Ambiguity in Multimodal Language Models]{Calibrated Ambiguity in Multimodal Language Models: \\Humans reach for cultural references, \\while models describe the picture}

\author{Cody Kommers}
\authornote{Both authors contributed equally to this research.}
\email{ckommers@turing.ac.uk}
\orcid{https://orcid.org/0009-0007-8985-0085}
\affiliation{%
  \institution{The Alan Turing Institute}
  \city{London}
  \country{United Kingdom}
}

\author{Mingrui Ye}
\authornotemark[1]
\email{mingrui.ye@kcl.ac.uk}
\orcid{https://orcid.org/0009-0002-8338-8778}
\affiliation{%
  \institution{King's College London}
  \city{London}
  \country{United Kingdom}
}

\author{Evelyn Gius}
\orcid{https://orcid.org/0000-0001-8888-8419}
\affiliation{%
  \institution{Technical University of Darmstadt}
  \city{Darmstadt}
  \country{Germany}
}

\author{Daniela Mihai}
\orcid{https://orcid.org/0000-0003-3368-9062}
\affiliation{%
  \institution{University of Southampton}
  \city{Southampton}
  \country{United Kingdom}
}

\author{Hoyt Long}
\orcid{https://orcid.org/0000-0002-8562-5426}
\affiliation{%
  \institution{University of Chicago}
  \city{Chicago}
  \country{United States of America}
}

\author{Zheng Yuan}
\orcid{https://orcid.org/0000-0003-2406-1708}
\affiliation{%
  \institution{University of Sheffield}
  \city{Sheffield}
  \country{United Kingdom}
}

\author{Drew Hemment}
\orcid{https://orcid.org/0000-0002-0068-5500}
\affiliation{%
  \institution{The Alan Turing Institute}
  \city{London}
  \country{United Kingdom}
}
\affiliation{%
  \institution{University of Edinburgh}
  \city{Edinburgh}
  \country{United Kingdom}
}

\renewcommand{\shortauthors}{Kommers, Ye et al.}

\begin{abstract}

Ambiguity is often treated as a bug for AI systems to resolve---but in human communication and culture, ambiguity can also be a generative resource. From humour to politics to art, people express themselves in words and images that are open enough to invite different interpretations, yet constrained enough to be interpretable. We operationalise this notion of \textit{calibrated ambiguity} with a task drawn from the parlour game \textit{Dixit}. We compare differences in clues generated by human vs multimodal language models, based on a novel coding rubric for calibrated ambiguity, and find that models consistently exhibit ambiguity collapse (i.e., their outputs are over-specified, leaving no room for multiple legitimate interpretations). Unlike human clues, AI-generated clues also exhibit cultural flattening; they almost never make reference to culturally-situated knowledge, even when prompted to use allusion and figurative language.

\end{abstract}

\ccsdesc[500]{Human-centered computing~Human computer interaction (HCI)}

\keywords{ambiguity, culture, calibration, homogenization, collapse, cultural flattening, multimodal language models, interpretive technologies}

\begin{document}

\maketitle

\begin{figure*}[t]
  \centering
 \includegraphics[width=\textwidth]{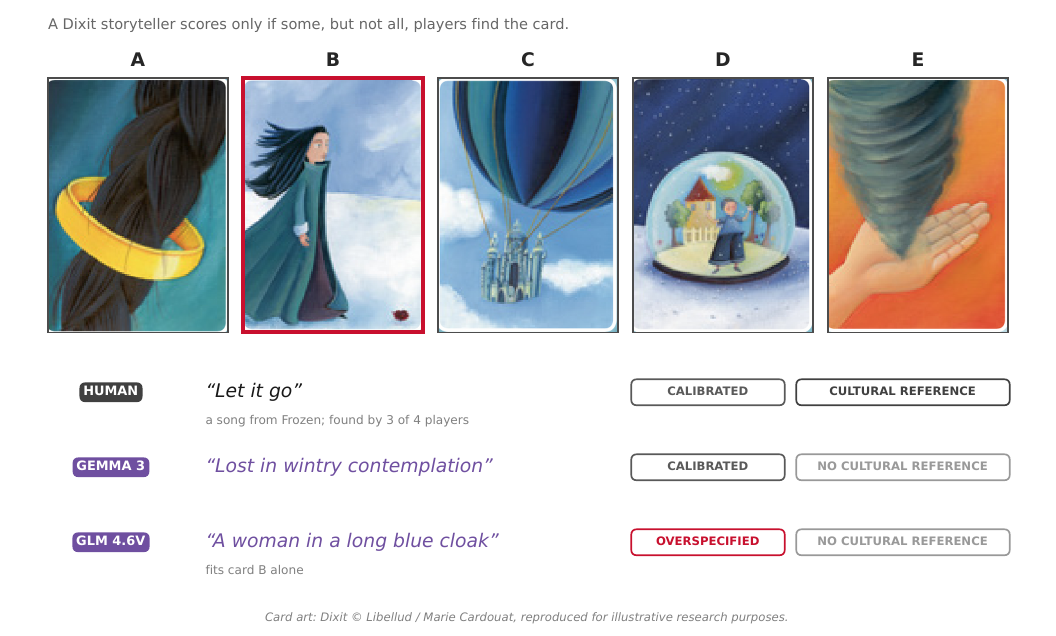}
 \caption{We use the parlour game Dixit as a paradigm for measuring calibrated ambiguity. The task requires a ``storyteller'' to generate a ``clue'' that ambiguously picks out a target card from a set of distractor---as judged by  being correctly selected by some but not all of the other players. In this example, a human player constructs their clue around an allusion to a song from the movie \textit{Frozen}---a widely-known cultural reference. Our results show how model-generated clues differ systematically from those given by humans. In this example, Gemma 3 offers a clue that is calibrated for ambiguity, but describes the image without reference to culturally-situated knowledge; while  GLM-4.6V simply provides an unambiguously literal clue.}
  \Description{Five Dixit cards labelled A to E, with card B, a cloaked figure in the snow, outlined in red as the target. Below, three clues for this board with rubric tags: the human clue Let it go, tagged calibrated and cultural reference; Gemma 3's clue Lost in wintry contemplation, tagged calibrated and no cultural reference; GLM 4.6V's clue A woman in a long blue cloak, tagged overspecified and no cultural reference.}
  \label{fig:teaser}
\end{figure*}

\input{text/01-introduction}
\input{text/02-related-work}
\input{text/03-method}
\input{text/04-coding-rubric}
\input{text/05-results}

\input{text/06-model-comparison}
\input{text/07-discussion}
\input{text/08-limitations}
\input{text/09-ethics}

\bibliographystyle{ACM-Reference-Format}
\bibliography{references}

\appendix
\input{text/appendix}

\end{document}

%% file: text/01-introduction.tex
\section{Introduction}

In AI research, ambiguity is often treated as a quantity to be minimised. A long tradition of work on word sense disambiguation~\cite{navigli2009wsd,yuan-strohmaier-2021-cambridge} and, more recently, benchmarks that test whether language models can resolve ambiguous questions and sentences~\cite{min2020ambigqa, liu2023ambient} share the premise that a system succeeds when ambiguity is removed. However, ambiguity is not just defined by semantic vagueness; it can be a vital aspect of human communication and culture~\cite{empson1930seven, piantadosi2012ambiguity,sep-ambiguity,kommers2026hermeneutics}.

For example, ambiguity plays a role in narrative, where explicit exposition of themes or insights tends to trivialise a story (e.g., ``telling'' rather than ``showing''~\cite{booth1961rhetoric}), and where the gaps a text leaves open are precisely what a reader must fill in to make it meaningful~\cite{iser1978act}. It plays a role in interpersonal conversation and pragmatic implicature, in which people gesture towards meaning rather than state it outright to avoid causing undue offence or tension~\cite{grice1975logic, brown1987politeness, pinker2008indirect}. Careful management of ambiguity can play a role in many deliberative processes: groups frequently reach agreement on what to do while leaving the underlying principles unresolved~\cite{sunstein1995incompletely}. When juries are faced with incomplete information, their collective job is to reach a conclusion even when ambiguity remains~\cite{hastie1983jury}.

Calibrating ambiguity is crucial to designing and evaluating generative AI as a cultural technology~\cite{farrell2025cultural, kommers2026hermeneutics}, as well as an important consideration for Interpretive Technologies: AI systems designed to navigate cultural complexity \cite{hemment2025doing}. AI systems are increasingly deployed for tasks that do not admit of straightforward ground-truth answers \cite{leibo2024appropriateness}. Evaluating the outputs of these systems requires not just binary judgments of correct and incorrect---but interpretive acts of the kind traditionally performed by humanities scholars \cite{klein2025provocations, underwood2025impact, hemment2025doing, beguvs2025artificial}. These interpretations account for ambiguity directly, specifically by accounting for a multiplicity of conflicting but legitimately held perspectives, rather than prioritising a canonical ground-truth answer \cite{kommers2026hermeneutics}. Accordingly, a significant risk of generative AI systems is ambiguity collapse \cite{gurarieh2026ambiguity}: whereas most real-world situations or outcomes can be described by multiple legitimate interpretations, AI interfaces often offer a single, authoritatively-presented judgment \cite{metzger2024empowering}. 

The challenge is that, by definition, something that is ambiguous resists a unitary ground-truth evaluation \cite{sep-ambiguity}. In its most general form, ambiguity simply means that a word, statement, image, or other cultural artefact can lend itself to multiple interpretations or meanings \cite{gurarieh2026ambiguity}. But there can be many shades of what this looks like. For example, Clint Eastwood's 1992 film \textit{Unforgiven} is characterised by moral ambiguity: each of the characters is trying to do the right thing, but ends up transgressing nonetheless. Who is the archetypal ``good guy'' the audience is supposed to root for? By contrast, Roy Lichtenstein's use of styles from graphic novels is also ambiguous. Are these works celebrating comic book art, or mocking it? There is no unified way to operationalise ambiguity comprehensively across even just these two cases---except to say that the artefacts in question admit of multiple interpretations. 

The calibration of ambiguity in AI systems is therefore fundamentally a design problem \cite{gaver2003ambiguity, di2025design}. We want AI systems that do more than just minimise ambiguity under any circumstances---but when, where, and how can ambiguity be constructively or usefully employed? The answer is context dependent: it depends on the people involved, their background, the situation they are in, and what they are trying to do \cite{kommers2026hermeneutics}. While there is no one-size-fits-all level of ambiguity that works in all cases for all people, appropriate management of ambiguity is especially important in settings where tasks are culturally-situated \cite{kommers2026hermeneutics, heuser2025cultural, veselovsky2026localized, de2026does, montesinos2026machine, yadav2025evaluation, liu2025culturally, bergman2023representation, sorensen2024roadmap, ge2024culture, roland2026social}. For example, AI systems are increasingly used to shape or produce stories and other other cultural artefacts \cite{gupta2026ai}. Such narratives provide a crucial structure for how people make sense of themselves and the world around them \cite{kommers2025sense, bruner1990acts}. At scale, how AI shapes narratives and cultural artefacts can therefore have a significant impact on identity and meaning-making \cite{kommers2026ai, kommers2026slop, zhi2026does}. But what it looks like for a cultural artefact (such as a story) to be appropriate within a given cultural frame depends not just on checking specific demographic boxes, but navigating and managing tradeoffs among intrinsically ambiguous concepts \cite{zhou2025culture}. Likewise, determining what kinds of stories count as meaningful to a given person or community depends not just on raw engagement metrics, but by accounting for the ambiguities inherent in contextually-situated perspectives \cite{kommers2025meaning, lowe2025full}. These considerations are notoriously difficult to quantify \cite{leibo2024appropriateness}. 

In this paper, we present a paradigm for measuring the calibration of ambiguity in multimodal AI systems. Whereas ``disambiguation'' has a clear operational definition~\cite{navigli2009wsd,yuan-strohmaier-2021-cambridge, liu2023ambient, tsai2026uncertain}, an evaluation framework for calibrated ambiguity is still needed \cite{gurarieh2026ambiguity}. We develop a task aimed at codifying the calibrated ambiguity and provide initial answers to the following questions: How well can generative AI systems calibrate ambiguity? What strategies do models rely on when managing ambiguity? How do these strategies differ from those of humans? And how might we calibrate ambiguity for a particular context or use-case?

\subsection{Summary of Contributions}

We present a novel methodology for measuring calibrated ambiguity in multimodal language models, based on the popular parlour game \emph{Dixit} (Section~\ref{sec:method}). We develop a novel annotation rubric to distinguish between different dimensions of ambiguity relevant to the task (Section~\ref{sec:rubric}). We find that humans and models tend to calibrate ambiguity in fundamentally different ways---with humans more likely to make use of culturally-situated knowledge (Section~\ref{sec:results}). In a model comparison, we show that there are significant differences between models in their default behaviour of how ambiguity is managed and show that prompting can alleviate miscalibration to a degree (specifically by emphasising a ``social cognition'' framing of the task), but cannot fully prevent ambiguity collapse (Section~\ref{sec:model-comparison}).

%% file: text/02-related-work.tex
\section{Related Work}
\label{sec:collapse-previous-work}

Previous work by Gur-Arieh, Wang, and Fazelpour~\cite{gurarieh2026ambiguity} establishes a taxonomy of epistemic risks posed by ambiguity collapse in large language models (LLMs)\footnote{We use 'LLM' rather than 'MLM' or 'VLM' in the text to reflect the common usage in this discussion, contextualising ambiguity collapse as a risk of generative AI systems broadly---though it should be remembered that the models we use are not only large, but multimodal.}; we explicitly aim to build on their work. Gur-Arieh et al.~\cite{gurarieh2026ambiguity} distinguish between three categories of risk associated with ambiguity collapse---process, output, and ecosystem. In this paper, we focus specifically on output collapse: the degree to which systems produce artefacts that support a plurality of legitimate interpretations. The concern about ambiguity collapse articulated by Gur-Arieh et al.~\cite{gurarieh2026ambiguity} reflects a larger concern about the homogenising force exerted by LLMs in collapsing or flattening culturally-situated outputs \cite{heuser2025cultural, xie2026artificial, xiao2025algorithmic, wilkens2026ai, jain2025task}.

\subsection{Measuring ambiguity}

While there is no established general measure for evaluating a computational system's capacity to calibrate ambiguity, previous work attempts to evaluate whether systems recognise or resolve multiple (reasonable) interpretations of an input \cite{min2020ambigqa, liu2023ambient, liu2023ambient, nam2025vague}. For example AmbigQA, for instance, evaluates whether question-answering systems recover answers corresponding to different interpretations of ambiguous questions \cite{min2020ambigqa}, while the benchmark \textsc{AmbiEnt} tests whether language models can represent alternative readings of ambiguous linguistic inputs \cite{liu2023ambient}. Wildenburg et al.~\cite{wildenburg2024dust} similarly provide a dataset for testing whether language models recognise semantic underspecification; Nam et al.~\cite{nam2025vague} extend ambiguity evaluation to multimodal models by examining whether visual context enables disambiguation; and Karim et al.~\cite{karim2025beyond} show how models can be calibrated for uncertainty. Previous work has also investigated how artists have utilised models as sources of ambiguity in their works \cite{tsai2026uncertain}.

However, few approaches position ambiguity as an explicitly desirable element of model output. The closest line of research is perhaps ``perspectivist'' annotation, premised on the observation that  in many datasets disagreement is desirable or expected rather than something to be eliminated \cite{basile2021we}. For example, recent work has taken this perspectivist approach to label disagreement in irony \cite{frenda2023epic} or offensive language \cite{kim2025analyzing}. However, ambiguity is only one potential cause of disagreement \cite{sandri2023don}; others include bias and noise \cite{uma2022scaling}, or misinterpretation and deficient definitions \cite{gius2017hermeneutic}. Other annotation work use variability as a proxy for ambiguity; for example, by using human annotation patterns to capture the appropriate amount of disagreement across interpretations \cite{dumitrache2019frame, nie2020collective, marchal2022establishing}. In this vein, Pavlick and Kwiatkowski~\cite{pavlick2019inherent} examine the shape of human judgments by testing for multiple modes within a distribution. Other work uses graded human ratings of ambiguity or indirectness to assess models' ability to integrate visual context for intent disambiguation \cite{nam2025vague} or uses semantic entropy to quantify uncertainty across semantically distinct model outputs \cite{farquhar2024semantic}. Together, this work gestures towards the way desirable variability, disagreement, or multiplicity has been evaluated in generative AI systems---specifically by using a human sample for reference in defining what form that variability ought to take.

\subsection{Dixit-based tasks}

In an effort to assess model performance on tasks requiring complex reasoning or which do not have a single right answer, researchers have increasingly turned to strategic games as a useful testbed \cite{lin2025gamebot}. One game that has attracted significant attention in this respect is the card game Dixit. Kunda and Rabkina~\cite{kunda2020creative} identified this task of ``creative captioning'' as a novel challenge for AI requiring the integration of visual perception, natural language understanding, and social reasoning. 

Existing computational work on Dixit has primarily treated the game as a paradigm for vision--language association. Iwata et al.~\cite{iwata2020dixit} developed an AI player to model the image--word associations involved in Dixit and found that the AI performed equally or worse than human players in clue generation, and worse in card selection and voting. Vatsakis et al.~\cite{vatsakis2022dixit} focused on the voting task, using human  data, natural language processing methods, and Internet search to predict the storyteller's card from a deliberately vague hint. Wei~\cite{wei2023dixit} subsequently approached card--hint matching with a neural network designed to learn visual concepts from natural language, while Chang~\cite{chang2024dixit} used the same neural network for hint selection. Finally, Tanzawa et al.~\cite{tanzawa2023hints} operationalise and quantitatively measure card-hint relevance, which they explicitly connect to the problem of producing appropriately ambiguous hints. They evaluate the relevance of hints for individual cards, comparing three human hint creators with one AI system, and find the lowest median relevance score and the greatest variability for the latter. 

These studies of Dixit have ranged widely in their scale and size. On one end, the controlled human evaluation studies are very small-scale---one 15-turn game \cite{iwata2020dixit} and the analysis of only 20 cards \cite{tanzawa2023hints}. At the other extreme, evaluations in Vatsakis et al.~\cite{vatsakis2022dixit}, Wei~\cite{wei2023dixit}, and Chang~\cite{chang2024dixit} rely on an archival dataset of approximately 116{,}000 human-played rounds of the game rather than newly collected
experimental judgments.

%% file: text/03-method.tex
\section{Method}
\label{sec:method}

\subsection{Dixit Paradigm}

Dixit is a parlour game designed around calibrated ambiguity. It is a multiplayer game featuring a deck of Dixit-specific cards, each with a surreal image of an object, person, or situation that could admit of a range of interpretations. Each round, one player acts as the ``storyteller'': they pick a card from their hand and offer a ``clue'' related to the image. Based on the clue, the other players pick an image from their own hands which they think could be described by the clue; all cards are then shuffled together, and the non-storyteller players vote for the card they believe was the storyteller's. Crucially, the storyteller wins the round only if \emph{some but not all} of the other players choose the image they put into the middle. The optimal clue is therefore one based on calibrated ambiguity: not so vague as to be uninterpretable, not so specific as to obviously map to the target image. 

\subsection{Existing dataset of human-generated clues}
\label{sec:dataset}

The dataset of human responses used in this study is based on the Dixit dataset introduced by Vatsakis et al.~\cite{vatsakis2022dixit} and made publicly available by the authors.\footnote{\url{https://www.spronck.net/datasets/Dixit_AI_data.zip}} The dataset was constructed from games played on the online platform \textit{boiteajeux.net} between July 2012 and September 2021 and is restricted to games using the 84 cards from the Dixit base game. As described by Vatsakis et al.~\cite{vatsakis2022dixit}, the original data were manually filtered to improve the quality and consistency of the textual information. Rounds in which the hint consisted solely of a web address were removed, as were rounds containing non-English hints. Exceptions were retained for commonly recognized foreign-language expressions (e.g., \textit{carpe diem}), titles of well-known works (e.g., \textit{Le Petit Prince}), and names referring to franchises, people, or landmarks. Web links appearing in storyteller explanations were also removed. Following this preprocessing, the resulting dataset contained 116{,}226 rounds with the storyteller's hint, the cards submitted during that round, the identity of the storyteller's card, and the votes received by each card. We use the dataset as released and did not alter its content; clue texts are reproduced verbatim, up to whitespace normalisation.

\paragraph{Integrity checks and study sample}

Before adopting the dataset we verified that the release matches its published description: it contains 116{,}226 rounds, and a storyteller's post-round explanation is present for 26\% of them. Because our boards require the five-player format (one target, four decoys, four guessers), we restricted the data to the 19{,}407 five-player rounds and checked the structural consistency of each: a round was retained only if its board comprised exactly five distinct cards from the base-game deck, exactly one card was marked as the storyteller's and agreed with the dataset's target field, and the vote record was complete, with votes summing to the four guessers. Only six rounds failed these checks, and every retained clue is English-language text, consistent with the authors' preprocessing. 
We selected a subset of 350 rounds sampled with fixed random seeds, excluding duplicate clue texts; this was used in as the set of boards (Section~\ref{sec:expanded}) for use in the various analyses detail in this paper. 

\subsection{LLM-generated Dixit Clues and Judgments}
\label{sec:llm}

We sought to compare human-generated clues with those generated by models. As
storytellers, vision--language models produced clues for the same cards that human storytellers had played, so that human- and model-generated clues can be compared on identical boards (Table~\ref{tab:models}). As judges, LLMs applied the ambiguity coding rubric of Section~\ref{sec:rubric} to clues, following the protocol applied by human coders.

\paragraph{Boards}
\label{sec:expanded}
All clues were generated for the same set of 350 five-player rounds drawn from the human dataset (see Section~\ref{sec:dataset} for selection criteria and integrity checks). Each round supplies the human storyteller's target card, the four decoy cards that the other players actually submitted, the human storyteller's own clue, and the votes it received. The set comprises 200 rounds sampled at random with a fixed seed (the \emph{baseline} boards) and the 150 rounds whose human clues were hand-coded in Section~\ref{sec:rubric}, so that every coded human clue has an LLM counterpart on the same board. 

\paragraph{Model selection}
\label{sec:model-selection}
We used six open-weight vision--language models as storytellers (Table~\ref{tab:models}). Three considerations guided the selection. First, every model is open-weight and was served locally in bfloat16 on a single 80\,GB GPU, so that generation ran under our own system prompt and sampling settings, with no API layer in between. Most models were roughly 30B total parameters, which is approximately the largest tier that fits one accelerator without quantization. Smaller models were not of interest (with the exception of GLM 4.6V Flash [9B], to have one example of a smaller model) because we wanted to focus systems that performed the task well, and larger ones could not be run under the same conditions. Second, the six models come from five developers and span both dense and mixture-of-experts (MoE) architectures. An MoE model routes each token through a few of many expert sub-networks~\cite{fedus2022switch}, and analyses of open MoE models find their experts more specialized and less polysemantic than dense feed-forward layers~\cite{lo2025closer}. Whether such internal differences surface in generation behaviour, for instance in how specific or how varied a clue is, has not been studied; the contrast lets us ask whether architecture predicts calibration (Section~\ref{sec:architecture}). Third, within these constraints the lineup is not size-matched: total parameters range from 9B to 35B and active parameters from 3B to 27B. GLM-4.6V-Flash (9B) is the only member of its family small enough to serve locally (the full GLM-4.6V has 106B parameters), and Kimi-VL is released at 16B only. We kept both because family diversity mattered more to us than exact size matching, and because size can then be examined across the lineup rather than held constant (Section~\ref{sec:architecture} finds that neither total nor active parameter count orders the models).

\begin{table}[t]
    \centering
    \small
    \caption{Storyteller models (``active'': parameters used per token in
    MoE models).}
    \label{tab:models}
    \begin{tabular}{@{}llll@{}}
        \toprule
        \textbf{Model} & \textbf{Developer} & \textbf{Architecture} & \textbf{Parameters} \\
        \midrule
        Gemma-3-27B-it        & Google   & dense & 27B \\
        Qwen3.8-27B           & Alibaba  & dense & 27B \\
        GLM-4.6V-Flash        & Zhipu AI & dense & 9B \\
        Qwen3.5-35B-A3B       & Alibaba  & MoE   & 35B (3B active) \\
        Kimi-VL-A3B-Instruct  & Moonshot & MoE   & 16B (3B active) \\
        ERNIE-4.5-VL-28B-A3B  & Baidu    & MoE   & 28B (3B active) \\
        \bottomrule
    \end{tabular}
\end{table}

\paragraph{Prompts}
Each model was run under two prompts. The \emph{minimal} prompt stated only the rules of the game and its scoring goal---that some but not all of the other players should identify the target---together with a length bound of three to six words and the instruction to respond with the clue alone. The \emph{social cognition} prompt added additional emphasis: it asked the model to consider how the other players will interpret the clue, to prefer figurative or abstract language that could apply to several cards, and to aim for a clue that exactly two of the four other players would resolve.

The contrast allowed us to ask whether explicitly prompting a model to reason about other minds changes the kind of ambiguity it produces \cite{strachan2024testing}. Because the social cognition prompt bundled several instructions---audience modelling, a preference for figurative or abstract language, and a numerical target---the comparison tested the prompt variations as a whole rather than any one component. Some of the clues coded by the human panel (Section~\ref{sec:human-v-llm-results}) were generated under an earlier wording of the social cognition prompt that differed only in its closing instruction; the panel rated the two wordings indistinguishably, and they are not distinguished below. Additional details of prompting protocols are available in Appendix~\ref{app:prompts}.

\paragraph{Clue generation}

Clues were sampled at temperature $0.9$ with a fixed random seed, with extended ``thinking'' disabled for models that support it, one clue per round for each model--prompt combination. The clue was extracted from the response by stripping markup and surrounding quotation marks. Clues were never truncated: if a generation exceeded the length bound, the model was resampled up to five times and the shortest complete attempt was kept, so that clue length is a property of the model rather than of post-processing. (Human clues were likewise unbounded in length; GLM-4.6V-Flash exceeds seven words in $17\%$ of rounds, against $10\%$ of human clues.) This yields $350 \times 2$ prompts $= 700$ clues per model and $4{,}200$ LLM clues in all.

To match the informational position of a human storyteller at the moment of clue-giving, a model saw \emph{only} its own target card---never the other players' cards---and was told that four other players would each contribute a decoy before all five cards were shuffled on the table. (Note: human storytellers chose which card to play, often with a clue already in
mind, whereas the models produced a clue for a card it did not choose)

\paragraph{Judgments: LLMs as rubric coders}

To compare producers at this scale we used LLMs to employ the coding rubric of Section~\ref{sec:rubric}, applied by LLM coders under the same protocol as the human annotators \cite{dunivin2025scaling, xiao2023supporting}. Each item was presented exactly as a human coder saw it: the clue and the five candidate card images (labelled A--E, in the order the human coders saw them), with the rubric as the system prompt. The coder first selected the image it believed to be the target and then rated the clue on all four dimensions. Each item was a single independent API call at temperature $0$, and the coder never saw the answer key, whether the clue was human- or LLM-generated, or the study hypotheses. We used two coders from different model generations, Gemini~3.1~Flash-Lite and Gemini~3.5~Flash-Lite. Every clue---the $4{,}200$ LLM clues and the 350 human clues on the same boards---was rated independently by both coders, giving $9{,}100$ ratings in total; unless stated otherwise, the analyses below use the mean of the two coders per clue.

Before scaling up, both coders were checked against the 150 clues that the five-annotator human panel coded (78 human and 72 LLM clues, Section~\ref{sec:rubric}). Per-clue coder ratings correlate with the human panel mean at $r = 0.60$--$0.79$ across the four dimensions (Calibration $0.62$--$0.68$, Situatedness $0.77$--$0.79$, Literalness $0.69$--$0.70$, Figurativeness $0.60$--$0.63$), which matches the agreement of an individual human annotator with the rest of the panel (leave-one-out $r = 0.58$--$0.80$ on the same dimensions). Exact agreement with the rounded panel mean ranges from $0.55$ (Figurativeness) to $0.80$ (Situatedness). LLM coders thus approximate the human panel about as well as individual human annotators do, on the subset the panel coded; on that basis we use their ratings for the model comparison of Section~\ref{sec:results-models} (full comparison in Section~\ref{sec:coder-validation}). 

An earlier version of the pipeline scored clues behaviourally, by having a panel of small vision--language models vote for the target card as a Dixit audience would. We retain it as a robustness check (Appendix~\ref{app:judges}); it reproduced the population-level human outcome but tracked individual clues only weakly; accordingly, we used the coding rubric as our primary instrument of assessment.

%% file: text/04-coding-rubric.tex
\section{Ambiguity Coding Rubric}
\label{sec:rubric}

We developed a novel coding rubric to account for the different ways of generating and managing ambiguity employed by humans and models when producing Dixit clues \cite{chinh2019ways}. The coding scheme feature four dimensions, each scored on a 3-point ordinal scale, which are summarised in Table~\ref{tab:rubric}. \emph{Calibration} asks whether the clue is overly specific, too vague, or just right. \emph{Situatedness} asks what degree of cultural knowledge is needed to interpret the clue. \emph{Literalness} and \emph{Figurativeness} ask, respectively, how literally and how figuratively the clue interprets the image; they are coded independently, since a single clue can be high on both (see Section~\ref{sec:iterations}). 

These dimensions were chosen to correspond directly to our hypotheses or considerations of interests, rather than to provide an exhaustive taxonomy of the different possible textures of ambiguity. Specifically, we wanted to investigate the following considerations: 

\begin{enumerate}
    \item \textit{A direct test of ambiguity collapse}. This was imperfectly captured by raw accuracy scores (i.e., do some but not all judges guess the correct card?). In pilot studies, we assessed ``collapse'' as it is done in the original game. The problem is that a lot of false positives occurred when the clue was completely ambiguous: guessing results in at least one judge getting the correct answer by chance. This is probably not a major issue when humans play the game in a social context. But it was insufficiently sensitive to detect when models simply failed to do the task. We therefore defined the calibration dimension as a more targeted instrument for assessing collapse.
    \item \textit{A direct test of cultural flattening}. In our informal initial surveying of human versus model clues, it was apparent that humans were making use of a cultural references in a way models were not. This is consistent with accounts of cultural collapse and homogenisation of model outputs \cite{heuser2025cultural, jain2025task}. We developed the situatedness dimension, using this term from the humanities \cite{kommers2026hermeneutics, haraway1988situated} to capture the degree to which specialised cultural knowledge was needed to interpret the clue.
    \item \textit{The degree to which a clue was constructed by reference to concrete visual features in the picture---or using more abstract, figurative devices}. Informally, it seemed that model-generated clues were more likely to pick out concrete visual features from an image. We explored several iterations on how best to capture this consideration (see Section~\ref{sec:iterations}).
\end{enumerate}

\begin{table}[t]
    \centering
    \small
    \caption{The ambiguity coding rubric. Each clue is rated on four
    dimensions, each on a 3-point ordinal scale.}
    \label{tab:rubric}
    \begin{tabular}{@{}>{\raggedright\arraybackslash}p{0.18\linewidth}
                       >{\raggedright\arraybackslash}p{0.24\linewidth}
                       >{\raggedright\arraybackslash}p{0.24\linewidth}
                       >{\raggedright\arraybackslash}p{0.24\linewidth}@{}}
        \toprule
        \textbf{Dimension} & \textbf{1} & \textbf{2} & \textbf{3} \\
        \midrule
        \textbf{Calibration} \newline \textit{Is the clue overly specific, too vague, or just right?}
          & \textbf{Underspecified} --- no confident inference to a single image is possible
          & \textbf{Calibrated} --- a plausible target is identifiable, but genuine doubt/alternative interpretations remain
          & \textbf{Overspecified} --- the clue deterministically picks out one image \\
        \addlinespace
        \textbf{Situatedness} \newline \textit{What degree of cultural knowledge is needed to interpret the clue?}
          & \textbf{Universal} --- interpretable via general English competence alone
          & \textbf{Mainstream} --- requires broadly-shared Anglophone/cross-cultural knowledge (e.g., major religious stories, globally-known figures)
          & \textbf{Niche} --- requires specific subcultural or specialised knowledge (e.g., a line from a specific film) \\
        \addlinespace
        \textbf{Literalness} \newline \textit{How literally does the clue interpret the image?}
          & \textbf{Low} --- no literal description; entirely figurative/abstract
          & \textbf{Mid/Mixed} --- some literal description, or unclear/debatable
          & \textbf{High} --- primarily literal description of depicted objects/features \\
        \addlinespace
        \textbf{Figurativeness} \newline \textit{How figuratively does the clue interpret the image?}
          & \textbf{Low} --- no figurative language; entirely literal
          & \textbf{Mid/Mixed} --- some figurative language, or unclear/debatable
          & \textbf{High} --- primarily a reference to figurative language, metaphor, allusion, etc. \\
        \bottomrule
    \end{tabular}
\end{table}

\subsection{Procedure}

Five of the authors of this paper acted as expert annotators. Before conducting the full set of annotations, a session was held with a majority of the annotators to review specific examples within the data set. The first 15 clues were discussed in depth, as well as several instructive examples that could be considered edge cases. The session was recorded and shared with annotators who were unable to attend.

Annotators viewed each clue in the context of a full 5-image set (target +
four distractors). They were blind to the answer key (i.e., which was the correct target image) as well as to whether the clue was human- or model-generated. Annotators first selected which image they believed to be the target, then rated each clue on all four dimensions.

A fifth rating was included, offering notes on any externally verifiable referents for each clue. For example, the clue ``Colors of the wind\ldots'' might be interpreted as being about literal, feature-based ambiguity---unless the interpreter knows that this is a reference to a song from the movie \textit{Pocahontas}. In this case, the clue does not refer to wind or colour, but to the depiction of a woman who (allegedly) resembles this film character. A single rater used a combination of Gemini and Google search to verify whether each clue had an external referent. Some were ambiguous, but many made clear reference to some specific cultural artefact. We decided it was better to try to control for this knowledge by noting these references, rather than to leave it up to the prior cultural knowledge of the raters.

\subsection{Rubric iterations}
\label{sec:iterations}

We went through several iterations of the rubric \cite{chinh2019ways}. In an initial version, we featured a category called ``mechanism,'' with four nominal labels distinguishing between whether the clue's ambiguity was primarily based in a feature, a concept, figurative language, or an indexical reference. However, many clues had multiple aspects of these. In a subsequent version, we developed one dimension meant to capture whether the ambiguous language was primarily literal (e.g., a vague description of visual features) versus figurative (e.g., an allusion to a cultural reference). However, many clues featured aspects of both. For example, consider the clue \textit{David and Goliath} intended to pick out a card with figures of two difference sizes. This is clearly figurative (rated 3/3). But the figurative language is used in reference to concrete visual features in the image; thus it also has a non-negligible degree of literalness as well (rated 2/2). It may have been possible to define an exact trade-off between the two, but it proved cognitively effortful for our raters to determine whether it a clue was primarily literal versus figurative. Accordingly, we settled on a final version distinguishing literalness and figurativeness as separate dimensions. 

\subsection{Reliability}

We calculated inter-rater reliability using Krippendorff's alpha (interval
metric, $n = 150$ clues, 5 raters) for each coding dimension: Calibration
($\alpha = 0.866$), Situatedness ($\alpha = 0.927$), Literalness
($\alpha = 0.884$), and Figurativeness ($\alpha = 0.859$). All four dimensions
exceeded conventional thresholds for good reliability ($\alpha \geq 0.80$).

%% file: text/05-results.tex
\section{Results}
\label{sec:results}

\begin{figure*}[t]
    \centering
    \includegraphics[width=\textwidth]{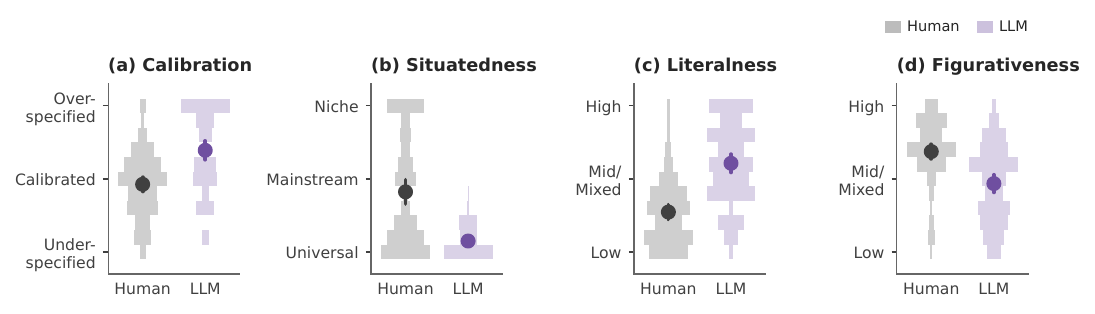}
    \caption{\textit{Human-coded rubric ratings on each of the four dimensions (Table~\ref{tab:rubric}) for human- vs. model-generated clues}. Shaded bars reflect the share of clues at each rating level; points indicate group mean with 95\% confidence interval. (a) Ambiguity collapse is demonstrated by models' calibration erring on the side of Over-specified. (b) Cultural flattening is shown by the tendency of models to generate clues that rely on universally situated knowledge (i.e., not requiring cultural knowledge for interpretation. (c,d) Models tend to be more literal and less figurative than humans by default.}
    \Description{Four panels, one per rubric dimension, each comparing human and LLM clues as stacked blocks of rating shares with a mean and confidence interval. LLM clues are rated more overspecified, far less culturally situated, more literal and less figurative than human clues.}
  \label{fig:main}
\end{figure*}

Our findings demonstrate key differences in how humans and LLMs calibrate ambiguity in the Dixit task (Fig. \ref{fig:main}). We investigated the degree to which models show ambiguity collapse---as well as how readily they draw on culturally situated knowledge in constructing their clues. We then analysed whether formal linguistic features are sufficient to explain these differences. In Section~\ref{sec:model-comparison}, we looked at the conditions under which various models differ in how they calibrate ambiguity.

\subsection{Humans and LLMs use different strategies to calibrate ambiguity---with varying levels of success}
\label{sec:human-v-llm-results}

For comparison of human- vs model-generated clues, a panel of five expert human annotators used our coding rubric to analyse a set of 150 clues mixed and blinded (78 human, 72 model), drawn from 150 rounds that form part of our 350 seed boards (Section~\ref{sec:expanded}). To models from our set of six (see Table~\ref{tab:models}) were used to generate clues for this comparison (35 Gemma 3 [27B], 37 Qwen 3.5 [35B]). The model clues were generated under the \emph{social cognition} prompt (Section~\ref{sec:llm}); the \emph{minimal} prompt of the expanded comparison is not represented in this set (Section~\ref{sec:prompting-mitigation}).

\subsubsection{LLMs exhibit ambiguity collapse}
\label{sec:llm-ambiguity-collapse}

 We sought explicit evidence of ambiguity output collapse in language models by coding human- vs model-generated clues according to ``calibration'' (as described in Section~\ref{sec:rubric}). Ambiguity output collapse was operationalized in this scheme by a rating of 3 on the Calibration dimension (where 2 is calibrated and 1 is overly ambiguous). 

In ratings from our five expert human annotators, LLM clues were rated as significantly more likely to be over-specified than human clues (mean calibration $2.39$ vs.\ $1.93$ on the 1--3 scale; Mann--Whitney $U$, $p < 0.001$; Fig.~\ref{fig:main}a). Humans tended to be calibrated for ambiguity. But on average, LLMs skewed toward being overly specific in their clues, thus exhibiting ambiguity output collapse.

\subsubsection{LLMs tend to flatten cultural context}
\label{sec:llm-cultural-flattening}

We investigated the degree to which models rely on culturally-situated knowledge when constructing their clues---such as allusions to music or movies, widely circulated idioms or stories, or idiosyncratic interpersonal knowledge (e.g., inside jokes or common acquaintances). As described in Section~\ref{sec:rubric}, our ``situatedness'' dimension encoded three levels: universal (no culturally situated knowledge), mainstream (broadly shared Anglophone knowledge), or niche (specialized cultural knowledge, e.g., a line from a particular film). 

In ratings from our five expert human annotators, LLM clues were rated as far less culturally situated than human clues (mean situatedness $1.15$ vs.\ $1.82$; $p < 0.001$; Fig.~\ref{fig:main}b). Human clues drew on a range of culturally specific references---which was almost never true of LLM clues. 

\subsubsection{LLMs tend to be more literal and less figurative than humans}
\label{sec:lit-vs-fig}

We also looked at the degree to which the clue offered a literal description of the image or relied on figurative language.
While these dimensions are strongly anti-correlated (Spearman's $\rho = -0.82$ across the $150$ human-coded clues, $p < 0.001$), it is possible for a clue-target mapping to rate high (or low) on both.

In ratings from our five expert human annotators, LLM clues were rated as significantly more likely to offer literal clues than humans (mean literalness $2.21$ vs.\ $1.55$; Mann--Whitney $U$, $p < 0.001$) and significantly less likely to rely on figurative language (mean figurativeness $1.94$ vs.\ $2.37$; $p < 0.001$; Fig.~\ref{fig:main}c--d). Some of the highly literal clues contributed to ambiguity collapse (i.e., increased incidence of over-specified clues), though much of this miscalibration reflected the models' failed attempts to generate ambiguity by providing vague descriptions of visual features rather than exact labels (e.g., using terms like ``wispy'' or ``fluffy'' for a target image featuring a cloud). 

\begin{figure*}[t]
    \centering
    \includegraphics[width=\textwidth]{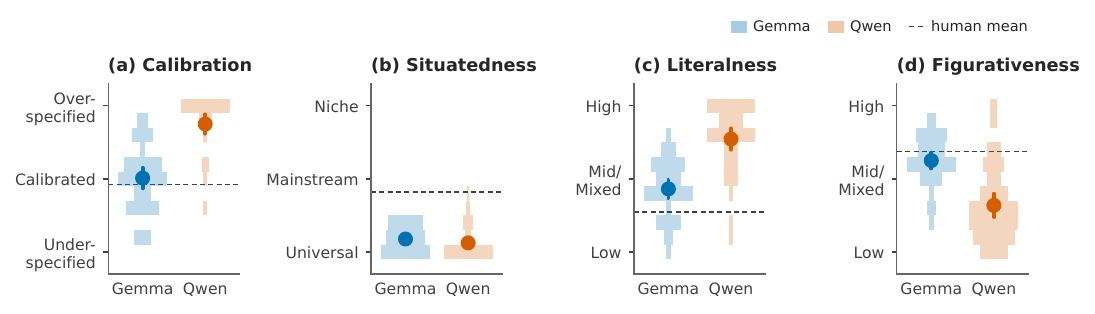}
    \caption{\textit{Model-generated clues from Fig.~\ref{fig:main} separated by model.} The dashed line reflects the the human mean. (a) Miscalibration is driven
primarily by Qwen 3.5 (35B), rather than Gemma 2 (27B). (b) Both avoid cultural references. (c,d) The preference for literal rather than figurative language is more pronounced in Qwen.}
    \Description{Four panels, one per rubric dimension, showing the LLM clues of the previous figure split into Gemma and Qwen, with the human mean as a dashed line. Both models sit at the floor of situatedness; Qwen is more overspecified and more literal than Gemma, whose calibration matches the human mean.}
  \label{fig:gemmaqwen}
\end{figure*}

\subsubsection{Ambiguity calibration depends crucially on the model}
\label{sec:initial_model_diff}

We used two different models to generate LLM clues for comparison with humans: Gemma 3 (27B; dense) and Qwen 3.5 (35B; mixture of experts; see Section~\ref{sec:llm}). Separating the outputs of these two models, we saw significant differences in the clues they generated (Fig.~\ref{fig:gemmaqwen}). Clues generated by both models were rated near the floor of the situatedness scale and did not differ significantly from one another (Gemma $1.18$, Qwen $1.12$; $p = 0.06$; Fig.~\ref{fig:gemmaqwen}b). However, they differed on the literalness and figurativeness dimensions: Qwen clues were more literal than Gemma clues (mean literalness $2.55$ vs.\ $1.86$; $p < 0.001$), while Gemma was more likely to draw on figurative language ($2.25$ vs.\ $1.64$; $p < 0.001$; Fig.~\ref{fig:gemmaqwen}c--d).

This gives an important qualification to our calibration claim: the effect of LLM clues as significantly over-specified compared to human clues (Section~\ref{sec:llm-ambiguity-collapse}) is driven by Qwen (mean calibration $2.75$ vs.\ $1.93$ for the human clues; $p < 0.001$), whereas Gemma's calibration is statistically indistinguishable from the human clues ($2.01$ vs.\ $1.93$; $p = 0.36$; Fig.~\ref{fig:gemmaqwen}a). Crucially, both models exhibit a similar degree of cultural flattening (Section~\ref{sec:llm-cultural-flattening}; situatedness $1.18$ and $1.12$ vs.\ $1.82$ for the human clues; both $p < 0.001$). In Section~\ref{sec:model-comparison}, we investigate whether this difference in calibration is an artefact of the models we initially selected or a robust difference in model behaviour.

\subsection{Can formal linguistic features capture the differences without the rubric?}
\label{sec:formal-features}

One potential concern about the findings presented in Section~\ref{sec:human-v-llm-results} is that the differences between human- and LLM-generated clues can be explained without reference to a complex, bespoke annotation rubric. Perhaps it would be enough to look at formal linguistic features to distinguish between these different strategies for calibrating ambiguity. Accordingly, we analysed whether a canonical set of formal linguistic features could predict whether a given clue was generated by a human or a model.

\subsubsection{Individually, formal features separate human and LLM clues only weakly}

We computed twelve surface features for every clue---length (word count, mean word length), lexical statistics (word frequency on the Zipf scale~\cite{vanheuven2014subtlex}, computed with the \textit{wordfreq} package~\cite{speer2022wordfreq}, and
concreteness~\cite{brysbaert2014concreteness}), part-of-speech composition (noun, adjective, determiner, and function-word ratios, presence of a finite verb), syntactic type, and orthography (title-casing, final punctuation)---for the 4{,}200 LLM-generated clues of the expanded clue set (six models $\times$ two prompts $\times$ 350 rounds; Section~\ref{sec:expanded}) and for a random sample of 2{,}000 human clues drawn from the 62{,}688 English clues in the full corpus. 

For each single feature we asked how well it alone would tell an LLM clue from a human clue. The area under the ROC curve (AUC) summarised this as the probability that a randomly chosen LLM clue scored higher on the feature than a randomly chosen human clue, with $0.50$ as chance and $1.0$ as perfect separation. With samples this large, small average differences are statistically reliable, so the confidence intervals of nearly every feature exclude $0.50$ (LLM clues are, on average, longer and more concrete). But statistical reliability does not translate directly to discriminative power: a difference in means can be significant while the two distributions have significant overlap. For example, $93\%$ of LLM clues had more than four words, while this was true of $46\%$ of human clues (Fig.~\ref{fig:surface}a, inset). Using word count to distinguish these clues would mark the majority of LLM clues correctly, but also misidentify almost half of human clues. Therefore, even the most discriminative formal feature on its own was not enough to distinguish human- and model-generated clues.

Word count was the most discriminative feature (AUC $= 0.74$, 95\% CI $[0.72, 0.75]$), followed by concreteness ($0.69$). The degree to which word count distinguishes between human and model clues is at least partially an artefact of our prompts (Appendix~\ref{app:prompts}). We asked models to generate clues in a length of 3-6 words. Humans featured more variation in their clues—some were a single word, while others were lengthy phrases. Concreteness maps well onto our literalness/figurative dimensions. This supports their inclusion as key dimensions in the rubric.

\subsubsection{LLM clues tend to be more homogeneous than human clues}
\label{sec:homogeneous-clusters}

\begin{figure*}[t]
    \centering
    \includegraphics[width=\textwidth]{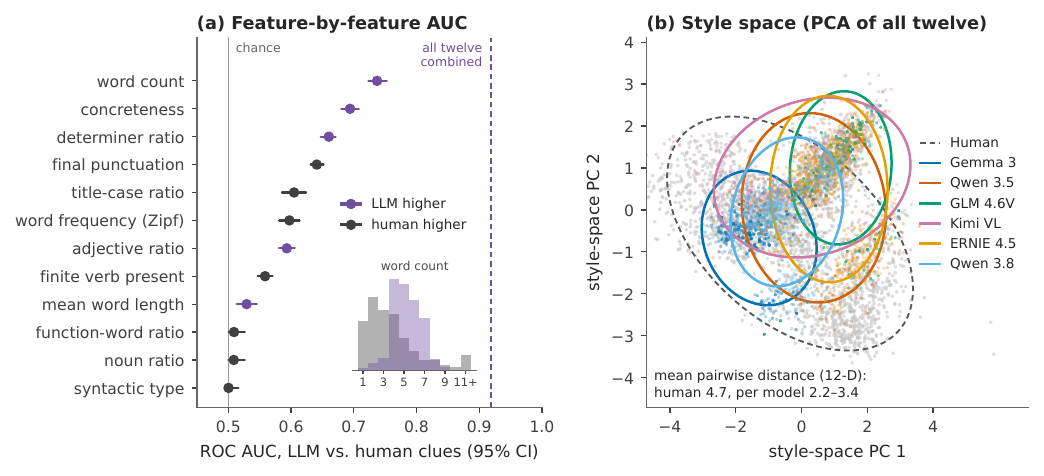}
    \caption{\textit{Formal linguistic features}. (a)~ROC AUC of each feature alone (95\% CIs); dashed line: all twelve combined (Mahalanobis score, held out); inset: word-count distributions. (b)~The same clues in the first two principal components; ellipses: 2\,SD.}
    \Description{Left: a dot plot of the ROC AUC of twelve surface features for telling LLM from human clues, all between the chance line at 0.5 and a dashed line at 0.92 for the combined twelve-feature score; word count is highest at 0.74, and an inset shows the overlapping word-count histograms. Right: a scatter plot of clues in two principal components, with a large dashed human ellipse and six smaller model ellipses lying inside it.}
  \label{fig:surface}
\end{figure*}

Taken as an aggregate, however, superficial features could identify at least one important systematic difference between human- and model-generated clues: homogeneity. In a one-to-one comparison of individual clues, the signal of any single formal feature was weak, but model-generated clues tended to look more like one another than human clues, which exhibit a much higher degree of variation. In short, LLM-generated clues tended to cluster more tightly than human-generated ones (Fig.~\ref{fig:surface}b). 

Specifically, a multivariate score (the Mahalanobis distance to the LLM centroid in the z-scored twelve-feature space) separated held-out LLM clues from human clues with AUC $= 0.92$ (the dashed line in Fig.~\ref{fig:surface}a), far above any single feature. In the twelve-feature space, LLM clues occupied a tighter subregion inside the human cloud (Fig.~\ref{fig:surface}b): $99.8\%$ of LLM clues fall inside the human 95\% ellipsoid, but only $23\%$ of human clues fall inside the LLM one. The mean pairwise distance between LLM clues was $3.2$ versus $4.7$ for human clues, roughly $1.5\times$ tighter. This aggregate figure potentially understates the effect: each model was narrower than the human cloud ($2.2$--$3.4$; $1.4$--$2.1\times$ tighter than humans). In the 2D projection (Fig.~\ref{fig:surface}b), the six models visually sit in different parts of the human cloud. The largest distances between model centroids ($3.2$ between Gemma and GLM; $0.7$--$3.2$ across all pairs) were twice the distance between the pooled LLM and human centroids ($1.6$). Thus, there was no single ``LLM style'', only a set of narrow, model-specific styles. Each of these styles could plausibly be generated by a human, but reliance on any single one dramatically underestimates the space of clues humans actually generate.

%% file: text/06-model-comparison.tex
\section{Model Comparison}
\label{sec:model-comparison}

In Section~\ref{sec:initial_model_diff}, our initial analysis suggested that different LLMs use different strategies for ambiguity calibration. This was supported by the analysis in Section~\ref{sec:homogeneous-clusters}, which showed that different LLMs produce clues characterised by separate clusters of formal features. To what degree are these just artefacts of the two models we chose to compare against humans? To test this, we ran a larger comparison of six models (Table~\ref{tab:models}), featuring a range of parameter sizes (9B--35B) and two different architectures (dense vs mixture of experts).

\begin{figure*}[t]
    \centering
    \includegraphics[width=\linewidth]{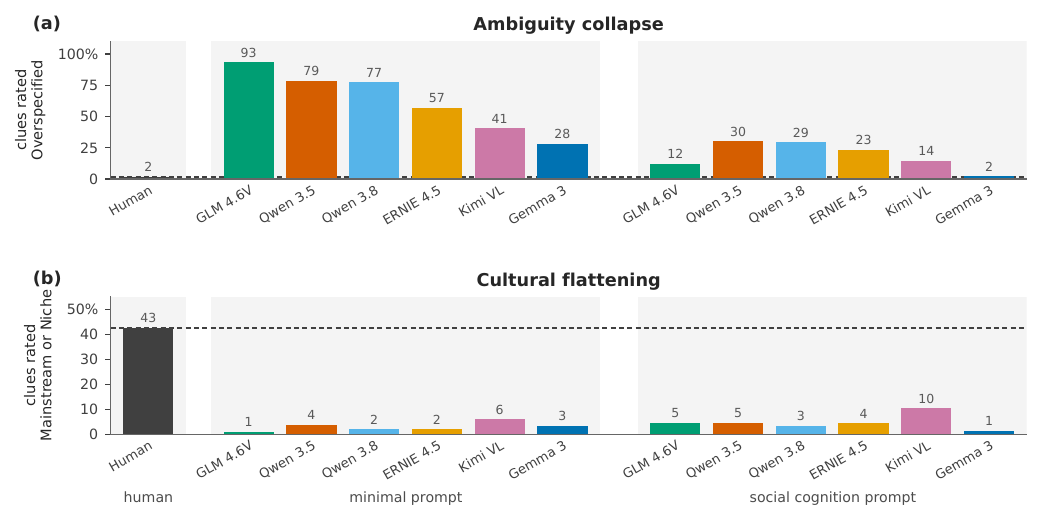}
    \caption{Models differ in their degree of ambiguity collapse, but all of them show evidence of cultural flattening. Each bank reflects analysis of 350 clues across six models under the minimal and the social cognition prompt, ordered by collapse under the minimal prompt. (a)~Share of clues both coders rated \emph{Overspecified}. (b)~Share rated \emph{Mainstream} or \emph{Niche} on situatedness; note the 50\% scale. Dashed lines: the human clues on the same rounds.}
    \Description{Two bar charts across thirteen groups: human clues, six models under the minimal prompt and the same six under the social cognition prompt. Top: share of clues rated overspecified, 2 percent for humans, 28 to 93 percent for models under the minimal prompt, 2 to 30 percent under the social cognition prompt. Bottom, on a 50 percent scale: share of clues rated culturally situated, 43 percent for humans and 1 to 10 percent for every model under either prompt.}
    \label{fig:models}
\end{figure*}

\subsection{Scaling annotation with LLM coders}
\label{sec:coder-validation}

To run this model comparison, we first needed to address the bottleneck of reliance on our human expert annotators. While there are recognized limitations to using LLMs for qualitative coding~\cite{xiao2023supporting, tai2024examination, ziems2024can}, a larger scale model comparison would not be feasible if it were limited by capacity of our human experts. 

To scale up our model comparison, we analysed how closely LLM coders could approximate the human coders when applying the rubric under the human protocol (Section~\ref{sec:rubric}).
On the 150 clues coded by the five human annotators, each LLM coder's rating correlated with the human panel mean at $r = 0.60$--$0.79$ depending on the dimension (Table~\ref{tab:coders})---close to, and on three of the four dimensions higher than, the agreement of an individual human annotator with the rest of the panel (leave-one-out $r = 0.58$--$0.80$). Exact agreement with the rounded panel mean was $0.55$--$0.80$, and the LLM coders identified the intended target card at least as often as the human annotators did ($0.62$--$0.65$ vs.\ $0.57$). Agreement was highest for situatedness, which was also the dimension on which the human annotators agree most with one another ($\alpha = 0.90$), and lowest for figurativeness and calibration, the two dimensions with the
lowest human reliability ($\alpha \approx 0.80$; Section~\ref{sec:rubric}).

\begin{table}[t]
    \centering
    \small
    \caption{LLM coders vs.\ the human panel ($n = 150$ clues). LOO $r$: one annotator vs.\ the mean of the other four. $r$, exact: correlation and exact agreement with the panel mean. Target: share of clues whose target card was identified.}
    \label{tab:coders}
    \resizebox{\columnwidth}{!}{
    \input{tables/coder_validation}
    }
\end{table}

On this subset the LLM coders also reached the same substantive conclusions. Re-running the comparisons of Section~\ref{sec:human-v-llm-results} with each LLM coder in place of the human panel reproduced the calibration result exactly---Gemma was indistinguishable from humans and Qwen was significantly overspecified (both coders: Human vs.\ Gemma n.s., Human vs.\ Qwen $p < 0.001$)---as well as the Human $<$ Gemma $<$ Qwen gradient in literalness and the collapse of situatedness for both models (all $p < 0.01$). 

\subsection{Model Comparison Results}
\label{sec:results-models}

After establishing broad agreement with the human annotators, we applied the two LLM coders to the full expanded clue set: six models $\times$ two prompts $\times$ 350 rounds $= 4{,}200$ LLM clues, plus the 350 human clues on the same boards, each rated independently by both coders ($9{,}100$ ratings). Ratings were averaged over the two coders per clue. Each of the twelve conditions (6 models $\times$ 2 prompts) were compared with the human clues on the same 350 rounds (Fig.~\ref{fig:models}; Table~\ref{tab:models-rubric}). This increased the number of clues being analysed, from $35$--$78$ clues per condition to $350$, and from two models to six. As described in Section~\ref{sec:llm}, we tested two different prompting conditions: minimal and social cognition. 

\subsubsection{All six models collapse, regardless of architecture or size}
\label{sec:all-model-collapse}

\begin{table*}[t]
    \centering
    \small
    \caption{Models differ in their degree of ambiguity collapse, but all of them show evidence of cultural flattening. Each bank reflects analysis of 350 clues across six models under the minimal and the social cognition prompt, ordered by collapse under the minimal prompt. (a) Share of clues both coders rated Overspecified. (b) Share rated Mainstream or Niche on situatedness; note the 50\% scale. Dashed lines: the human clues on the same rounds.}
    \label{tab:models-rubric}
    \input{tables/model_comparison}
\end{table*}

All six models exhibited ambiguity collapse under the minimal prompt. Each model produced clues that were rated as substantially more specified than the human clues on the same boards (all $p < 10^{-40}$; Cliff's $\delta = 0.54$--$0.97$; Fig.~\ref{fig:models}a). The model-specific collapse we observed in Section~\ref{sec:human-v-llm-results} is therefore not just an effect of Qwen. 

However, in some models the effects were more pronounced than others (Fig.~\ref{fig:models}). GLM-4.6V-Flash was rated as over-specified on $93\%$ of rounds ($2\%$ for the human clues). At the other end, Gemma-3-27B was labeled as over-specified on $28\%$ of rounds and Kimi-VL on $41\%$. This accorded with the coders' target-card choices: when shown a minimal-prompt model clue, they identified the target on $83$--$100\%$ of boards, against $34\%$ for the human clues on the same boards (Table~\ref{tab:models-rubric}, target id.). Clue length did not explain the difference: the collapsed banks range from $3.5$ to $6.4$ words per clue around the human mean of $4.0$. (Note: Section~\ref{sec:initial_model_diff} found no miscalibration for Gemma, but the human-coded set contained no minimal-prompt clues; under the social cognition prompt Gemma's gap remains negligible here as well, Section~\ref{sec:prompting-mitigation}.)

\label{sec:architecture}
We designed this comparison to contrast dense and mixture-of-experts models within a broadly similar scale (9B--35B total parameters; Table~\ref{tab:models}). Within this sample, calibration did not align with architecture. The two Qwen models---Qwen3.5-35B-A3B, a mixture-of-experts model
with 3B active parameters, and Qwen3.8-27B, a dense model of a different size and generation---are almost indistinguishable on every dimension under both prompts (calibration $2.87$ vs.\ $2.86$ under the minimal prompt and $2.42$ vs.\ $2.43$ under social cognition; literalness $2.87$ vs.\ $2.87$ and $2.30$ vs.\ $2.37$; figurativeness $1.19$ vs.\ $1.18$ and $2.10$ vs.\ $1.97$). 

Meanwhile the three dense models spanned a wider range, from the best-calibrated (Gemma) to the worst (GLM); the same was true of the three mixture-of-experts models. Neither parameter count nor active parameter count predicted the models' ranking. 

\subsubsection{Prompting mitigates collapse but does not remove it}
\label{sec:prompting-mitigation}

The social cognition prompt substantially changed the models' calibration. A two-way analysis of variance on the LLM clues yielded significant effects of prompt ($F(1, 4188) = 1590$), model ($F(5, 4188) = 145$) and their interaction ($F(5, 4188) = 37.9$; all $p < 0.001$) on calibration. The social cognition prompt influenced models to look more like the human distribution---though the size of the difference depended on the model. GLM shifted by $0.76$ points on the three-point scale (from $2.96$ to $2.21$), the two Qwen models by $0.43$--$0.45$, Gemma by $0.35$, ERNIE and Kimi by $0.32$. In short: the social cognition prompt was most helpful for models that exhibit the sharpest initial degree of collapse.

\begin{table*}[t]
    \centering
    \small
    \caption{Example clues from all 6 model and 2 prompt permutations compared with humans for two boards, shown with their five cards (target in red). Cal.: coder-averaged calibration rating (1 Underspecified, 2 Calibrated, 3 Overspecified). min/soc: minimal / social cognition prompt.}
    \label{tab:examples}
    \begin{tabular}{@{}llp{4.6cm}cp{4.6cm}c@{}}
        \toprule
        & & \multicolumn{2}{c}{\textbf{Board 1} (3 of 4 players found the target)}
          & \multicolumn{2}{c}{\textbf{Board 2} (1 of 4)} \\
        & & \multicolumn{2}{c}{\includegraphics[width=5.3cm]{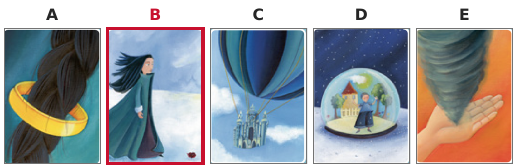}}
          & \multicolumn{2}{c}{\includegraphics[width=5.3cm]{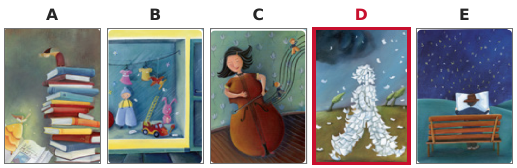}} \\
        \cmidrule(lr){3-4}\cmidrule(lr){5-6}
        \textbf{Producer} & & \textbf{Clue} & \textbf{Cal.} & \textbf{Clue} & \textbf{Cal.} \\
        \midrule
        Human & & Let it go & 2.0 & Christina Rossetti & 2.0 \\
        \midrule
        GLM-4.6V-Flash & min & A woman in a long blue cloak & 3.0 & A white figure in a stormy field & 3.0 \\
        Qwen3.5-35B & min & Girl in green coat in snow & 3.0 & Snowman built by a child & 2.5 \\
        Qwen3.8-27B & min & Long hair blowing in wind & 3.0 & Walking through scattered white paper & 3.0 \\
        ERNIE-4.5-VL & min & A figure in a long cloak & 3.0 & A ghost made of paper & 3.0 \\
        Kimi-VL-A3B & min & A woman with a head injury & 2.0 & I left my shirt in your car & 2.0 \\
        Gemma-3-27B & min & Lost in wintry contemplation & 2.0 & Ephemeral, fleeting, almost gone & 2.0 \\
        \midrule
        GLM-4.6V-Flash & soc & The lone wanderer in the vast, silent expanse & 2.0 & The ghost of a forgotten dream & 2.0 \\
        Qwen3.5-35B & soc & Cold winds of change are coming & 2.0 & Walking with a heart made of paper & 2.0 \\
        Qwen3.8-27B & soc & Walking through frozen tears & 2.0 & Walking on fragile thoughts & 2.0 \\
        ERNIE-4.5-VL & soc & A lonely wanderer in a long coat & 3.0 & A phantom in a tear-stained dream & 2.0 \\
        Kimi-VL-A3B & soc & a beacon of hope amidst despair & 2.0 & The heart of winter & 2.0 \\
        Gemma-3-27B & soc & Lost in quiet contemplation & 2.0 & Fleeting, fragile, and hopeful & 2.0 \\
        \bottomrule
    \end{tabular}
\end{table*}

Only Gemma reached a level of ambiguity calibration consistent with human-generated clues. Its social cognition clues were rated \emph{Calibrated} on $98\%$ of rounds ($97\%$ for the human clues) and received exactly the same rating as the human clue on $76\%$ of boards; the residual difference was statistically detectable at $n = 350$ ($p < 0.001$) but negligible in size ($\delta = 0.10$; with the Gemini 3.5 coder alone, $\Delta = 0.02$, $p = 0.20$). For every other model the gap remained substantial ($\delta = 0.25$--$0.53$; more specified than the human clue on $31$--$57\%$ of boards).

\label{sec:describe-the-pic}
The social cognition prompt asked the model to prefer figurative, rather than literal, language. For the most part, models were able to oblige. Under that prompt the figurativeness of Gemma's and GLM's clues ($2.55$ and $2.61$) actually exceeded that of the human clues ($2.42$; $p < 0.01$), and Kimi's and ERNIE's match it ($2.32$--$2.33$, n.s.; Table~\ref{tab:models-rubric}). Yet GLM, Kimi and ERNIE remain significantly overspecified: GLM's social-cognition clues, the most figurative bank in the study, were still rated \emph{Overspecified} on $12\%$ of rounds and more specified than the human clue on $31\%$ of boards. Therefore, prompting is an important lever for modulating the degree to which they ``describe the picture,'' as we claim in our title---though it is crucial note that figurative language did not translate directly to better calibration of ambiguity.

\subsubsection{Cultural flattening persists across models and prompts}
\label{sec:avoid-cultural-knowledge}

Our initial findings on cultural flattening in model-based clues (Section~\ref{sec:llm-cultural-flattening}) were supported by this analysis. Models uniformly relied on the ``universal'' level of situatedness, requiring no situated cultural knowledge to interpret their target (Fig.~\ref{fig:models}b). Human clues on these boards were rated ``mainstream'' or ``niche'' $43\%$ of the time ($11\%$ niche). The LLM clues were rated above \emph{Universal} on $3.9\%$ of rounds. Every one of the twelve conditions (6 models $\times$ 2 prompts) showed near-floor effects on the scale ($1.01$--$1.07$; $\delta = -0.34$ to $-0.42$ against the human clues, all $p < 0.001$). Unlike calibration, situatedness did not respond to the prompt (prompt effect $F(1, 4188) = 3.3$, $p = 0.07$; interaction $p = 0.24$); the only model effect is Kimi-VL's marginally higher floor. Collapse and cultural flattening therefore have different profiles in our sample. Collapse varies by model and is partly reversed by the social cognition prompt, whereas cultural flattening is present across models and is not modulated by prompt.

%% file: tables/coder_validation.tex
\begin{tabular}{@{}lccccc@{}}
\toprule
& \textbf{Human} & \multicolumn{2}{c}{\textbf{Gemini 3.1 Flash-Lite}} & \multicolumn{2}{c}{\textbf{Gemini 3.5 Flash-Lite}} \\
\cmidrule(lr){3-4} \cmidrule(lr){5-6}
\textbf{Dimension} & LOO $r$ & $r$ & exact & $r$ & exact \\
\midrule
Calibration & 0.60 & 0.68 & 0.77 & 0.62 & 0.76 \\
Situatedness & 0.80 & 0.77 & 0.80 & 0.79 & 0.78 \\
Literalness & 0.66 & 0.70 & 0.67 & 0.69 & 0.64 \\
Figurativeness & 0.58 & 0.63 & 0.60 & 0.60 & 0.55 \\
\midrule
Target identified & 0.57 & \multicolumn{2}{c}{0.65} & \multicolumn{2}{c}{0.62} \\
\bottomrule
\end{tabular}

%% file: tables/model_comparison.tex
\begin{tabular}{@{}llcccccc@{}}
\toprule
\textbf{Model} & \textbf{Prompt} & \textbf{Calib.} & \textbf{Situat.} & \textbf{Literal.} & \textbf{Figur.} & \textbf{Target id.} & \textbf{Words} \\
\midrule
Human & --- & 2.01 & 1.46 & 1.61 & 2.42 & 0.34 & 4.0 \\
\midrule
GLM-4.6V-Flash & min & 2.96 (+0.97) & 1.01 (-0.42) & 2.99 (+0.98) & 1.03 (-0.94) & 1.00 & 6.4 \\
 & soc & 2.21 (+0.27) & 1.02 (-0.39) & 1.96 (+0.37) & 2.61 (+0.22) & 0.83 & 6.4 \\
Qwen3.5-35B & min & 2.87 (+0.93) & 1.03 (-0.40) & 2.87 (+0.94) & 1.19 (-0.87) & 0.99 & 5.2 \\
 & soc & 2.42 (+0.51) & 1.03 (-0.39) & 2.30 (+0.64) & 2.10 (-0.26) & 0.92 & 6.1 \\
Qwen3.8-27B & min & 2.86 (+0.92) & 1.01 (-0.41) & 2.87 (+0.95) & 1.18 (-0.87) & 0.99 & 4.6 \\
 & soc & 2.43 (+0.53) & 1.02 (-0.40) & 2.37 (+0.70) & 1.97 (-0.36) & 0.95 & 4.7 \\
ERNIE-4.5-VL & min & 2.70 (+0.79) & 1.01 (-0.41) & 2.70 (+0.87) & 1.41 (-0.75) & 0.95 & 5.5 \\
 & soc & 2.38 (+0.49) & 1.02 (-0.40) & 2.23 (+0.62) & 2.33 (-0.06) & 0.93 & 6.5 \\
Kimi-VL-A3B & min & 2.52 (+0.61) & 1.05 (-0.37) & 2.50 (+0.73) & 1.48 (-0.70) & 0.83 & 3.5 \\
 & soc & 2.21 (+0.25) & 1.07 (-0.34) & 2.02 (+0.40) & 2.32 (-0.06) & 0.72 & 5.1 \\
Gemma-3-27B & min & 2.43 (+0.54) & 1.02 (-0.40) & 2.34 (+0.69) & 1.94 (-0.43) & 0.97 & 4.2 \\
 & soc & 2.07 (+0.10) & 1.01 (-0.42) & 1.85 (+0.30) & 2.55 (+0.12) & 0.87 & 4.1 \\
\bottomrule
\end{tabular}

%% file: text/07-discussion.tex
\section{Discussion}
\label{sec:discussion}

We present a novel paradigm for operationalising calibrated ambiguity in multimodal generative AI systems. Across several analyses, we compare clues from an online data set of human-generated Dixit clues with clues generated by vision–language models playing the same role on the same boards. We developed a coding rubric to distinguish key differences in how ambiguity is calibrated and managed by humans and models. Our main finding is that models, as compared with humans, tend to be less well calibrated and almost never rely on the kind of cultural references found in human-constructed clues. In follow-up analyses, we show that individual surface-level features distinguish human vs model clues only weakly, and that there is significant variation in how different models calibrate ambiguity.

\subsection{Do LLMs exhibit ambiguity output collapse?}

Yes. In our paradigm, models tend to be less calibrated for ambiguity than humans (Section~\ref{sec:human-v-llm-results}). This ``output collapse'' identified in our findings is only a subset of possible forms of ambiguity collapse articulated by Gur-Arieh et al.~\cite{gurarieh2026ambiguity}---but an important one that has been hypothesised, yet not demonstrated, in previous work. 

There are caveats to this claim. The prompt makes a difference (Section~\ref{sec:all-model-collapse} and Section~\ref{sec:prompting-mitigation}), though it does not mitigate ambiguity collapse entirely. A range of model sizes (9B--35B) and model architectures (dense, mixture-of-experts) are susceptible to this form of ambiguity collapse. However, different models do not collapse to the same degree. For example, under a ``social cognition'' prompt, Gemma 3 (27B) exhibited human-level ambiguity calibration (Section~\ref{sec:prompting-mitigation}). But overall, we found robust evidence for ambiguity collapse across a range of circumstances.

\subsection{Do LLMs avoid cultural references when calibrating ambiguity?}

To a significant degree, yes. In our paradigm, models avoided culturally situated knowledge (Section~\ref{sec:llm-cultural-flattening}). This effect held even under the social cognition prompt, which invites allusion and figurative language (Section~\ref{sec:prompting-mitigation}). This seemed to be one of the defining differences between how humans and LLMs managed ambiguity in our task (Section~\ref{sec:human-v-llm-results}). As a default behaviour, models tended to construct clues without cultural references. 

This is a significant capability gap in model outputs, consistent with the output-based risks described by Gur-Arieh~\cite{gurarieh2026ambiguity} and the hermeneutic challenges offered by Kommers et al.~\cite{kommers2026hermeneutics}. The evidence from our results suggests that culturally-situated knowledge plays a key role in how humans generate, manage, and calibrate ambiguity. If models avoid it by default, they are missing out an important communicative resource. However, it remains an open question how best to elicit cultural information from models, for example by prompting them with information about the audience for whom they were generating clues or offering implicit clues about context without directly stating it (e.g., using British spellings) \cite{veselovsky2026localized}. Nonetheless, humans naturally reference popular movies or stories like Pocahontas and Frozen; they riff on cultural idioms and stock phrases; and they play on associations with single words or concepts (like Australia or Metallica). These rely on cultural knowledge widely shared in the Anglophone world---and depending on the clue-target pairing in question, little expertise is required for interpretation beyond an awareness that such cultural referent exists. This is consistent with emerging evidence that LLMs struggle to ``fill in the blanks'' with details about hypothetical personas~\cite{wang2025misportray}.

\subsection{Do models tend to calibrate ambiguity in the same way?}

No, not necessarily. The dynamics of ambiguity calibration seem to be model-specific (Section~\ref{sec:initial_model_diff}). For example, models differed in their proclivity for literal vs figurative language (Section~\ref{sec:lit-vs-fig}). We found that models tended to produce homogeneous clues, but with each model using an idiosyncratic strategy that differed from other models (Fig.~\ref{fig:surface}; Section~\ref{sec:formal-features}). Models also responded differently to different prompts (Section~\ref{sec:prompting-mitigation}). The one effect that was entirely robust across conditions and models was an avoidance of clues based on culturally-situated knowledge (Section~\ref{sec:avoid-cultural-knowledge}); models were consistently and uniformly averse to making cultural references in their clues.

\subsection{Can ambiguity collapse be fixed with prompting?}

A little bit, sometimes, and potentially. We provided models with a ``minimal'' prompt (just the rules of the game), and this resulted in unanimous ambiguity collapse (Section~\ref{sec:all-model-collapse}). The least calibrated models performed better when we adjusted the prompt to emphasise social cognition (Section~\ref{sec:prompting-mitigation}), but this did not by any means obviate the effect of collapse. Again, there were model-based differences. Of the models we looked at, Gemma performed best---and in some cases at human-level performance (Section~\ref{sec:prompting-mitigation}). 

In short, the evidence is mixed. It seems that prompting can help some models alleviate miscalibration to some degree. Targeted prompts can encourage models to use specific strategies (e.g., more figurative language)---but this does not automatically make them ``good'' at using these strategies in the sense that they result in more calibrated clues (Section~\ref{sec:describe-the-pic}). Thus, we stand by our primary claim that when calibrating ambiguity humans tend to reach for cultural references, while models tend to describe the picture.

%% file: text/08-limitations.tex
\section{Limitations and Alternative Perspectives}
\label{sec:limitations}

\subsection{I agree the ambiguity matters. But does your task actually measure the kind of ambiguity we care about?}

The strength of this task is that it provides a measurable, operationalised definition of calibrated ambiguity, where none existed before. The cost of that precision is that it only gets at a limited range of the possible textures of ambiguity. We operationalise ambiguity by the presence of multiple legitimate interpretations. This is kind of like operationalising a joke by whether it makes someone laugh. It is relevant---but it is also far from the whole story. The three non-calibration dimensions of our rubric (situatedness, literalness, and figurativeness) are meant to be broadly applicable to different types of ambiguity, but the mapping is imperfect. For example, in the introduction we give the example of moral ambiguity in the film \textit{Unforgiven}. Certainly the dimension of culturally-situated knowledge would be relevant in analysing this form of ambiguity (what counts as morally good or bad behaviour in a particular frame of reference?). But it is less clear whether the tension between literal and figurative language or depictions is especially important. Our position is that the paradigm we present in this paper is a useful baseline. It captures some (not all) important aspects of ambiguity that apply in a range of situates. But ultimately the true ``calibration'' of ambiguity is tied intimately to the context in which a system is deployed. In future work, variations on this paradigm could be used to investigate more specific kinds of ambiguity (e.g., moral ambiguity in narratives produced by AI based on a set of Dixit-like target images) or variations could be adapted for specific use cases with a specific target population in mind (e.g., calibrating the ambiguity in labelling inappropriate images for content moderation for a given platform or population).

\subsection{You developed a paradigm for measuring calibrated ambiguity. But once we have identified miscalibration, how do we fix it?}

Our findings suggest that prompting makes a difference, especially for overall calibration and collapse. Of course it does, and the number of permutations for exploring how this might work are unbounded. Again, the key will be in contextualisation. Our paradigm offers a general framework for measuring calibrated ambiguity---but the kind of ambiguity that is desirable will depend on a specific situation or use-case. Future work should explore how prompting methods can achieve specific kinds of ambiguity---for specific models, operating in specific contexts. Our findings about the effects of prompting on cultural flattening are less clear. The relevant question is not whether models can be coaxed into drawing on cultural references, but about the processes for doing so that yield the most compelling results. There is a tension in existing literature between evidence for widespread homogenisation and cultural collapse \cite{jain2025task,heuser2025cultural} versus evidence that local cultural knowledge is controllable, if you know how to elicit it \cite{veselovsky2026localized}. Do models produce homogeneous outputs because we are just not prompting them correctly? It is a difficult question to answer, but the evidence from Veselovsky et al.~\cite{veselovsky2026localized} suggests that the key is not just asking behaviour representing a particular cultural milieu---but actively providing cues of participation in it. This is consistent with the view of LLMs as ``context machines'' \cite{kommers2026hermeneutics} which are fundamentally designed to adapt to the contextual cues with which they are provided. Future work can explore the transition from measuring calibrat\textit{ed} ambiguity to best practices for calibrat\textit{ing} it.

\subsection{I suspect these effects would be obviated in bigger models. Do you really think Claude Fable 5 would exhibit ambiguity collapse?}

This is an important direction for future research. Our primary aim in this paper was to provide a proof of concept for a novel methodology of measuring ambiguity calibration in LLM outputs. With this foundation, there are a lot of further questions to pose and investigate. How ambiguity calibration scales with model size is definitely one of them. In this version, we chose to use models that were large enough to do the task (e.g., Gemma 3 [27B] performed near human level in some cases) but lightweight enough to deploy in a range of situations (see Section~\ref{sec:model-selection} for further rationale). An additional part of the consideration was finding the right match for variation in performance for a simple baseline version of the task. For example, the task can be made arbitrarily more difficult by increasing the number of distractors. Given that Gemma 3 (27B) performs well under the conditions we studied, it is probably the case that bigger, more sophisticated models could do the task as currently designed. The task can be adapted for larger models---for example, by increasing the number of distractors, altering the kind of cards that are used, or providing background information about the shared cultural frames relevant for the other players in the game. Future work can explore how variations of this paradigm can elucidate the strategies more sophisticated models use to calibrate ambiguity---and provide a concrete measure of when they collapse.

%% file: text/09-ethics.tex
\section*{Ethics and Privacy Statement}
\label{sec:ethics}
All data for human-generated clues in this study were publicly available. All expert coders were authors on the paper, not participants in a study. Dixit images are subject to copyright (c) Libellud / Marie Cardouat. The work presented in our system presents minimal ethical or privacy risk. The Dixit images do not feature harmful, triggering, or inappropriate content. However, an adaptive version of this study could---specifically one that looks at ambiguity in a more ethically fraught context, such as content moderation.

%% file: text/appendix.tex
\section{Storyteller prompts}
\label{app:prompts}

Both prompts were given as the system message together with the target
card image. Text in the two prompts is identical up to the point marked
``\ldots''; the social cognition prompt then adds the coaching shown.

\paragraph{Minimal}
\begin{quote}\small
You are playing a game with four other players. In the game, you secretly
hold a TARGET card. Your job is to be the storyteller: you must generate a
clue that hints at your TARGET. The other four players each hold a hand of
cards and, after hearing your clue, will each submit one of their own
cards as a decoy that might fit your clue. All five cards (your TARGET plus
their four decoys) will then be shuffled on the table, and each of them
will vote for which one is the TARGET. You want some of the other players
to guess your TARGET correctly, but not all of them.

Your clue should be three to six words in length. Respond with only the
clue.
\end{quote}

\paragraph{Social cognition}
\begin{quote}\small
You are playing a game with four other players. In the game, you secretly
hold a TARGET card. Your job is to be the storyteller: you must generate a
clue that hints at your TARGET. However, you don't want the answer to be
obvious. The other four players \ldots{} You want some of the other players
to guess your TARGET correctly, but not all of them. When generating the
clue, think about things that could plausibly apply to many cards in a way
that might mislead some players, while still favoring the TARGET.

An ideal clue should be:
\begin{itemize}\setlength{\itemsep}{0pt}
    \item Three to six words in length
    \item Based on emotions, metaphors, allusions, and other forms of
          figurative or abstract language that could be applied to
          multiple cards
    \item Ambiguous enough that not all players will agree on the same
          card, but more likely to refer to your TARGET than to an
          unrelated card
\end{itemize}

Think about how the other players will perceive your clue: Respond with
only a clue that two of the four other players would correctly guess
applies to the TARGET rather than to a decoy.
\end{quote}

\section{Behavioural robustness check: a vision--language judge panel}
\label{app:judges}

Before adopting the rubric as the primary instrument, we scored clues
behaviourally with a panel of judges that voted for the target card as a
Dixit audience would. This check used an earlier configuration of the
pipeline: 200 four-player rounds from the human dataset, Gemma-3-27B and
Qwen3.5-35B-A3B as storytellers under the two wordings of the social cognition prompt ($200 \times 2 \times 2 = 800$ clues), with clues
clipped to a maximum of seven words.

\paragraph{Judge panel}
The panel comprised four open-weight vision--language judges (Qwen3.5-9B,
Qwen3-VL-8B, Gemma-3-12B, and InternVL3-8B), each served independently.
For every clue, each judge saw the four table cards from the original
round---the target plus the three decoys the human players had actually
contributed---in a shuffled order together with the clue, and voted for
the single card it believed the clue referred to (greedy decoding, forced
single-index reply). Votes that could not be parsed were discarded; the
full four-judge panel returned valid votes in 76\% of the 1{,}000 judged
rounds, and all metrics are computed over the valid votes of each round.
Valid-vote counts are essentially identical across producer conditions
(means of 3.60--3.69 judges per round), so discarded votes do not favour
any producer. Following Dixit's scoring rule, a clue counts as \emph{in
the target range} when some but not all valid judges select the target
card; clues that every judge resolves are overspecified (``too obvious''),
and clues that no judge resolves are underspecified (``too obscure''). The
same judge panel scored human- and LLM-generated clues on the same boards,
so any difference between producers cannot be attributed to the judging
procedure. The four judges behave as a heterogeneous audience rather than
four copies of one opinion: mean pairwise agreement on the voted card is
$0.45$ (chance $0.25$), and the panel is unanimous in only $14\%$ of
full-panel rounds.

\paragraph{Validation against human audiences}
The 200 human-clue rounds come with the real audience's outcome, which
lets us ask how well the judge panel stands in for human guessers. At the
population level the two audiences agree closely: the panel places
$60.5\%$ of human clues in the target range, against $60.0\%$ for the
original human audiences on the same rounds. At the level of individual
rounds the correspondence is weaker: per-round hit rates correlate
positively but modestly (Spearman $\rho = .29$, $p < .001$), and binary
in-range agreement between the two audiences is close to chance. Some
attenuation is inevitable when comparing single samples of two small
audiences (three human guessers vs.\ four model judges), but it does not
account for all of the gap: a binomial simulation in which both audiences
respond to the same per-clue transparency predicts substantially higher
correspondence (agreement $\approx .61$, $\rho \approx .53$) than we
observe. The panel is therefore not a stand-in for how a particular human
audience reads a particular clue, and we use it only for what its
aggregate behaviour is shown to match: population-level comparisons of
clue producers on identical boards.

\subsection{Model Comparison Robustness}

Three checks indicate that these results do not depend on incidental
features of the pipeline. First, the two coders agree with each other on
the $4{,}550$ double-coded clues at $r = .66$ (Calibration), $.73$
(Situatedness), $.75$ (Literalness) and $.84$ (Figurativeness), with exact
agreement of $.76$--$.96$; every calibration gap reported above is
significant in the same direction with either coder alone, the sole
exception being the Gemma social-cognition bank, which one coder rates as
equivalent to the human clues. Second, the 350 boards comprise two subsets
selected in different ways (Section~\ref{sec:expanded}); the mean
calibration of every bank differs by at most $0.10$ between them. Third,
length does not explain calibration. Within the LLM clues the correlation
between word count and calibration rating is $\rho = .11$; Kimi-VL's
minimal-prompt clues are the shortest of any bank ($3.5$ words) yet
substantially overspecified ($\delta = .61$), and GLM's two banks have
identical length ($6.4$ words) but $\delta = .97$ and $.27$. Length, in
other words, does not account for the calibration gap.